\documentclass[10pt,twocolumn,letterpaper]{article}

\usepackage{iccv}              

\definecolor{iccvblue}{rgb}{0.21,0.49,0.74}
\usepackage[pagebackref,breaklinks,colorlinks,allcolors=iccvblue]{hyperref}
\usepackage{acro}
\usepackage{multirow}

\DeclareAcronym{mAP}{short = mAP, long = mean Average Precision}
\DeclareAcronym{IoU}{short = IoU, long = Intersection over Union}
\DeclareAcronym{tal}{short = TAL, long = temporal action localization}
\DeclareAcronym{tsm}{short = TSM, long = Temporal Shift Module}

\def\paperID{*****} 
\def\confName{ICCV}
\def\confYear{2025}

\title{Multi-task Learning with Extended Temporal Shift Module for Temporal Action Localization}

\author{Anh-Kiet Duong\\
L3i Laboratory, La Rochelle University\\
17042 La Rochelle Cedex 1 - France\\
{\tt\small anh.duong@univ-lr.fr}
\and
Petra Gomez-Krämer\\
L3i Laboratory, La Rochelle University\\
17042 La Rochelle Cedex 1 - France\\
{\tt\small petra.gomez@univ-lr.fr}
}

\begin{document}
\maketitle
\begin{abstract}
We present our solution to the BinEgo-360 Challenge at ICCV 2025, which focuses on temporal action localization (TAL) in multi-perspective and multi-modal video settings. The challenge provides a dataset containing panoramic, third-person, and egocentric recordings, annotated with fine-grained action classes. Our approach is built on the Temporal Shift Module (TSM), which we extend to handle TAL by introducing a background class and classifying fixed-length non-overlapping intervals. We employ a multi-task learning framework that jointly optimizes for scene classification and TAL, leveraging contextual cues between actions and environments. Finally, we integrate multiple models through a weighted ensemble strategy, which improves robustness and consistency of predictions. Our method is ranked first in both the initial and extended rounds of the competition, demonstrating the effectiveness of combining multi-task learning, an efficient backbone, and ensemble learning for TAL.
\end{abstract}    
\section{Introduction}
\label{sec:intro}
Understanding human actions in complex real-world environments is a central problem in computer vision, with direct applications in robotics, augmented/virtual reality, and human-centric video intelligence. Traditional approaches to action recognition often rely on single-view visual data, which can be limited by occlusion, restricted fields of view, and the absence of contextual cues. For instance, egocentric views often capture the actor’s immediate focus, while third-person exocentric views provide global context but miss fine-grained interaction details. Such limitations motivate research on multi-perspective and multi-modal video analysis, where complementary information is jointly leveraged to achieve more robust and holistic understanding~\cite{sigurdsson2016hollywood}.

To address these challenges, the BinEgo-360 Challenge at ICCV 2025 introduces a new benchmark for \ac{tal} in multi-perspective and multi-modal settings. Unlike prior \ac{tal} datasets such as THUMOS~\cite{THUMOS14}, ActivityNet~\cite{caba2015activitynet}, or HACS~\cite{zhao2019hacs}, which primarily rely on monocular third-person video, this challenge incorporates diverse modalities: 360° panoramic video, third-person frontal video, egocentric monocular and binocular video, spatial audio, GPS and weather metadata, and textual scene-level descriptions. In addition to \ac{tal}, the challenge also features a complementary classification track, where the goal is to predict high-level scene categories from the same set of multi-view, multi-modal inputs. By combining egocentric and exocentric perspectives with auditory and environmental cues, the challenge provides a unique opportunity to explore richer fusion strategies and to advance beyond conventional single-stream pipelines.

The task is defined as detecting the start and end time of every action instance inside a video clip, along with its corresponding category label. Evaluation follows a standardized protocol based on \ac{mAP} across multiple temporal \ac{IoU} thresholds. This evaluation emphasizes both semantic correctness and temporal precision, reflecting real-world requirements where intelligent systems must not only recognize what action occurs but also localize exactly when it happens \cite{lin2019bmn}.

Overall, the BinEgo-360 Challenge establishes a new testbed for investigating how multi-view and multi-modal cues can be effectively combined for temporal action localization. Beyond benchmarking, it aims to foster the development of models that can generalize across heterogeneous environments, pushing the frontier of video understanding toward practical deployment in robotics, augmented/virtual reality, and human-centric perception \cite{li2024mvbench}.

\section{Related work}
\label{sec:related}

This section reviews prior work that is most relevant to our approach. We divide the discussion into two parts: video classification, which focuses on recognizing high-level activities or scene categories, and temporal action localization, which further requires detecting the start and end times of action instances within untrimmed videos.

\subsection{Video classification}

Video classification has been a long-standing problem in computer vision, aiming to recognize high-level activities or scene categories from untrimmed clips. Early works relied on hand-crafted features and two-stream architectures that process RGB frames and optical flow separately~\cite{simonyan2014two}. With the advent of deep learning, 3D convolutional networks such as C3D~\cite{tran2015learning} and I3D~\cite{carreira2017quo} were introduced to jointly capture spatial and temporal information. More recent approaches have focused on efficient temporal modeling, including SlowFast networks~\cite{feichtenhofer2019slowfast} and the \ac{tsm}~\cite{lin2019tsm}, which achieve strong performance with lower computational cost. Transformer-based architectures such as ViViT~\cite{arnab2021vivit} and TimeSformer~\cite{bertasius2021space} further extend these ideas by directly modeling long-range temporal dependencies with self-attention.

Progress in video classification has been driven by large-scale benchmarks, including Sports-1M~\cite{karpathy2014large}, Kinetics~\cite{kay2017kinetics}, and Something-Something~\cite{goyal2017something}, which emphasize diverse environments and fine-grained human-object interactions. Scene-level classification has also benefited from datasets such as Places~\cite{zhou2017places}, which provide rich context for indoor and outdoor categories. Together, these datasets and methods have established the foundations for scene classification tasks in multi-modal video understanding.

\subsection{Temporal action localization}

Temporal action localization extends action recognition by requiring not only the correct class label but also the start and end times of each action instance. Early methods often relied on sliding-window proposals and classification networks~\cite{shou2016temporal, yuan2016temporal}. Later anchor-based models such as SSN and TAL-Net introduced structured temporal anchors and refined boundary estimation~\cite{zhao2017temporal, chao2018rethinking}. More recent approaches focus on anchor-free paradigms, where temporal boundaries are directly regressed, as in Boundary-Matching Networks (BMN)~\cite{lin2019bmn} and Boundary Content Graph (BCG)~\cite{bai2020boundary}. Transformer-based frameworks have also been explored to capture long-range dependencies and contextual cues in untrimmed videos~\cite{xu2020g, liu2022end}.

Several large-scale datasets have played a critical role in advancing \ac{tal}. THUMOS14~\cite{THUMOS14} and ActivityNet~\cite{caba2015activitynet} remain standard benchmarks for temporal detection, providing densely annotated untrimmed videos across diverse action classes. HACS~\cite{zhao2019hacs} further scales up with human action clips and segments, while EPIC-Kitchens~\cite{damen2018scaling} introduces egocentric recordings that emphasize daily activities in unconstrained environments. These benchmarks highlight challenges such as dense action labeling, long-tail distributions, and domain generalization, and continue to drive progress in both algorithm design and evaluation.

\section{Methodology}
\label{sec:method}

\begin{figure*}
    \centering
    \includegraphics[width=0.95\textwidth]{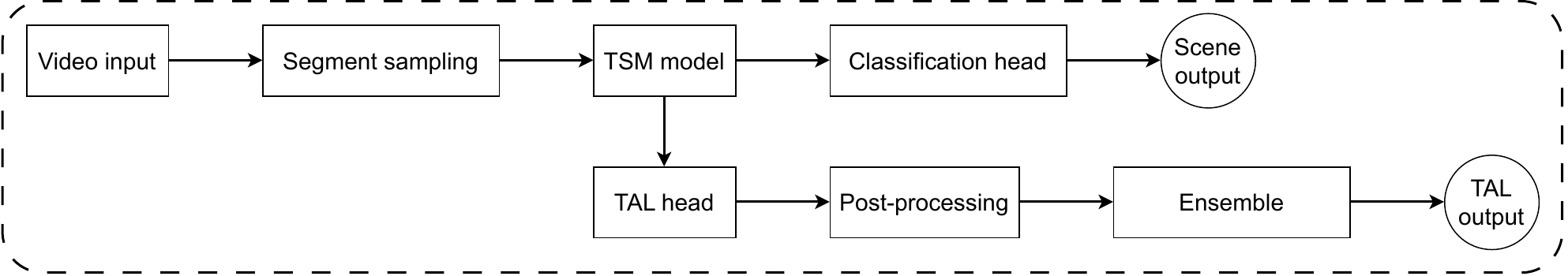}
    \caption{Illustration of extending TSM for temporal localization.}
    \label{fig:method_full}
\end{figure*}

This section describes our proposed method, which is composed of three main components. We first present a multi-task learning framework that jointly addresses scene classification and temporal action localization. We then explain how the Temporal Shift Module (TSM) is extended to support localization by predicting actions in fixed-length intervals with a background class. Finally, we introduce an ensemble strategy to combine multiple models for more robust predictions. An overview of the entire framework is illustrated in Figure~\ref{fig:method_full}.

\subsection{Multi-task learning}
The BinEgo-360 Challenge defines two tasks: scene classification and \ac{tal}. The classification task requires predicting the scene category of a video clip, ranging from indoor (\eg kitchen, bars, office) to outdoor (\eg park, street, nature). In contrast, TAL aims to detect both the action label and its temporal boundaries within an untrimmed video. Although the objectives differ, the two tasks are closely related. For instance, actions such as eating or ordering food are more likely to occur in dining or food outlets, while cooking actions are typically associated with a kitchen scene. Leveraging such dependencies can improve overall model performance.

To exploit this connection, we adopt a multi-task learning framework in which a shared backbone is trained jointly for both classification and \ac{tal}. As the backbone, we choose the \ac{tsm}~\cite{lin2019tsm}, a state-of-the-art architecture for video understanding. \ac{tsm} captures temporal dynamics through lightweight shift operations across channels, offering high accuracy with relatively low computational cost. Beyond standard benchmarks, \ac{tsm} has also achieved strong results in action recognition challenges~\cite{duong2024action}, highlighting its robustness and generalization ability across diverse video understanding tasks.

This makes \ac{tsm} a natural starting point for our approach, serving as the foundation for the multi-task learning framework described in the following subsections.

\subsection{Extending TSM for temporal localization}

\begin{figure}
    \centering
    \includegraphics[width=0.95\linewidth]{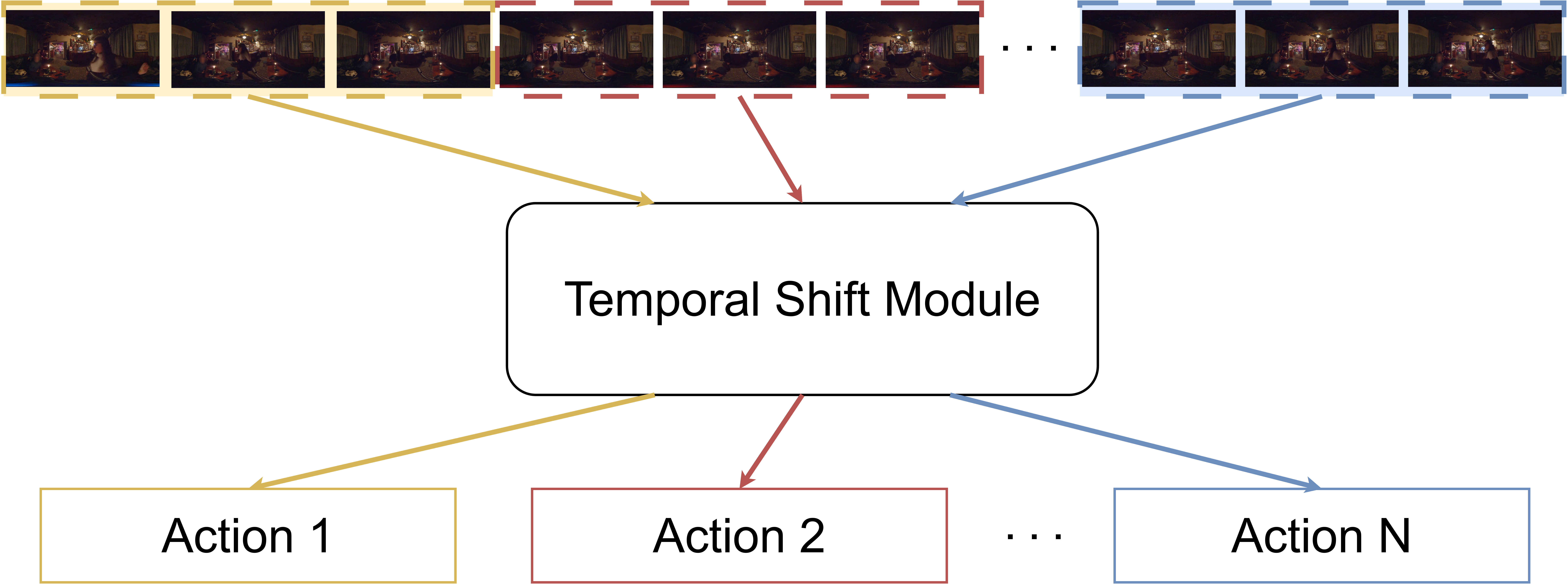}
    \caption{Illustration of extending TSM for temporal localization.}
    \label{fig:ext_tsm}
\end{figure}

We begin by introducing an additional label to represent background segments where no action occurs. The TSM is then trained on the dataset with $N_{\text{class}}+1$ categories, where $N_{\text{class}}$ is the number of annotated actions and the extra class corresponds to the no-action label. 

At inference time, let $L$ denote the length of a video and $t$ a predefined interval size. The video is partitioned into $\lfloor L/t \rfloor$ consecutive non-overlapping intervals, and the trained TSM is used to classify each interval into one of the $N_{\text{class}}+1$ categories. 

To improve temporal consistency, we apply a post-processing step where consecutive intervals assigned to the same action label are merged into a longer segment, with the confidence score set to the maximum among them. This merging reduces fragmentation and produces cleaner action intervals.

This extension of TSM retains the low computational complexity of the original method while benefiting from its strong classification performance, which supports the multi-task learning framework. However, it also has limitations, such as the risk of missing very short actions or failing to capture multiple actions occurring simultaneously. \Cref{fig:ext_tsm} summarizes this extension, and to address its limitations we further introduce ensemble learning in the following section.

\subsection{Ensemble learning}

Ensemble methods have long played an important role in machine learning competitions, where the combination of multiple models often leads to more stable and higher-ranking solutions. By aggregating predictions from diverse models, ensemble approaches reduce the risk of overfitting to specific data patterns and improve robustness to noise and uncertainty. Beyond challenges, ensemble learning has also been widely adopted in real-world systems, where the ability to balance complementary strengths of different models is critical for achieving consistent performance across heterogeneous environments.

In this work, we implement a weighted ensemble of several TAL models. Each submission file contains a set of predictions formatted as $(class, start, end, confidence)$. We first parse all submission files and align them by video identifier. For each video, we then create a dictionary of candidate segments, indexed by their class and temporal boundaries $(class, start, end)$. Since the same segment may be predicted by different models with different confidence values, we maintain a list of confidences across models for each candidate segment. To combine them, we compute a weighted average:
\[
\hat{c} = \frac{\sum_i w_i \cdot c_i}{\sum_i w_i},
\]
where $c_i$ is the confidence from model $i$ and $w_i$ is its assigned weight. The weights are chosen to reflect the relative reliability of each model, based on validation performance. Segments that appear in multiple models thus receive higher confidence if they are consistently supported across models.

After aggregating scores, we apply a post-processing step to merge overlapping segments of the same class. Specifically, for two segments, we compute the temporal Intersection-over-Union (IoU). If the IoU exceeds a threshold, we merge them by expanding the boundaries to cover the union of both intervals and keep the maximum confidence. This step consolidates redundant detections and reduces noise from small temporal variations among models. The final output is a single prediction file that integrates the strengths of all models, resulting in more reliable and robust temporal localization.

\section{Experiments}
\label{sec:experiments}

In this section, we present the experimental evaluation of our approach. We first describe the dataset used for training and evaluation, followed by the experimental setup including implementation details. We then report the main results of the competition. Finally, we conduct ablation studies to analyze the contribution of different components of our method.

\subsection{Dataset}

\begin{figure}
    \centering
    \subfloat[Panoramic]{\includegraphics[width=0.45\linewidth]{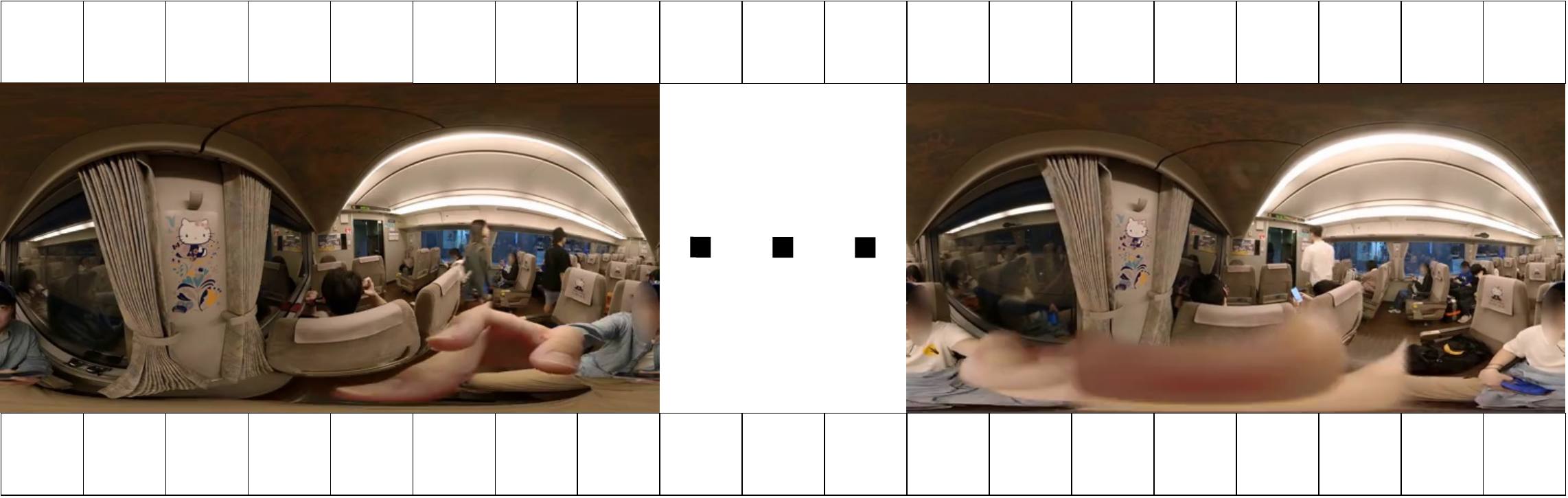}} 
    \quad 
    \subfloat[Third person]{\includegraphics[width=0.45\linewidth]{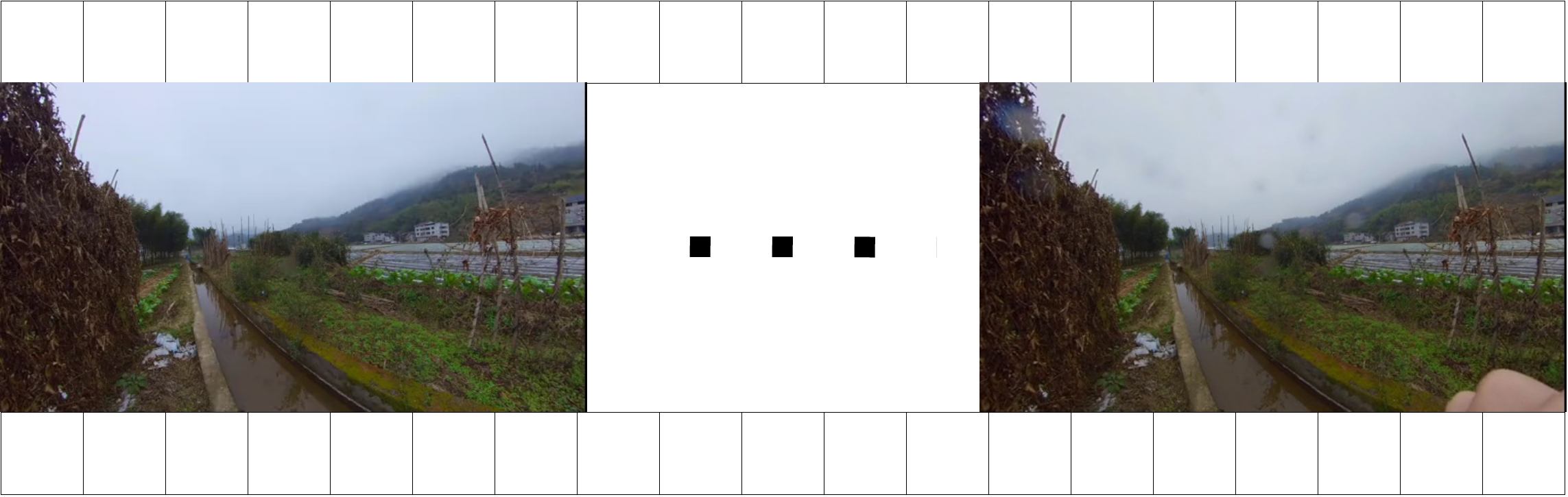}}
    
    \subfloat[Monocular]{\includegraphics[width=0.45\linewidth]{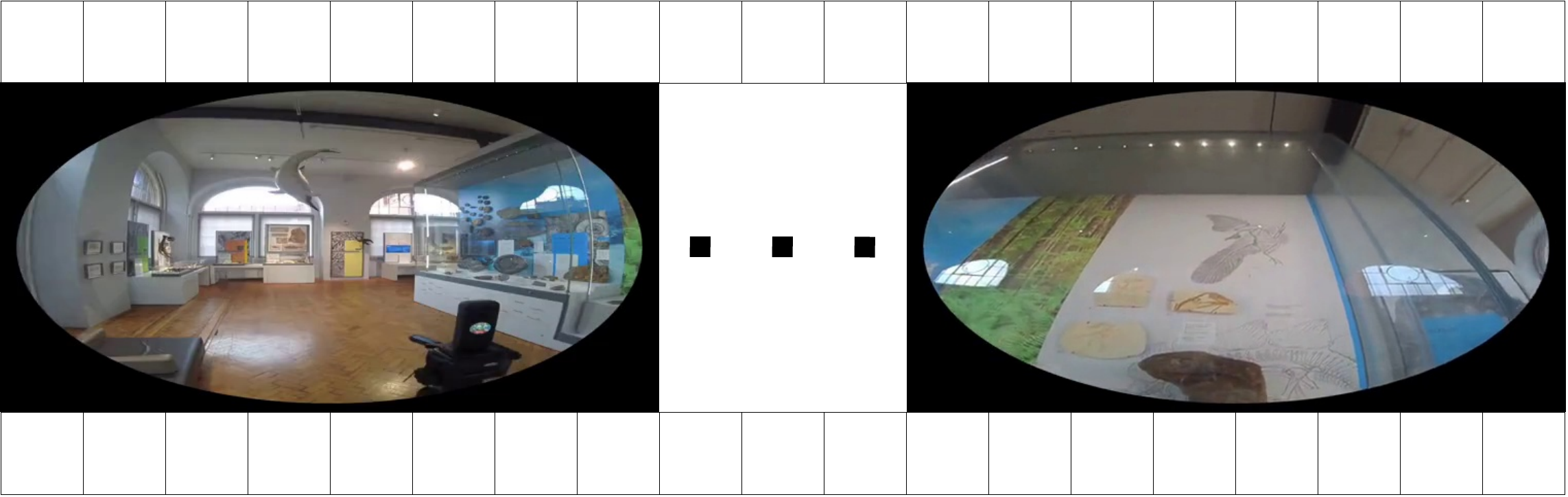}} 
    \quad 
    \subfloat[Binocular]{\includegraphics[width=0.45\linewidth]{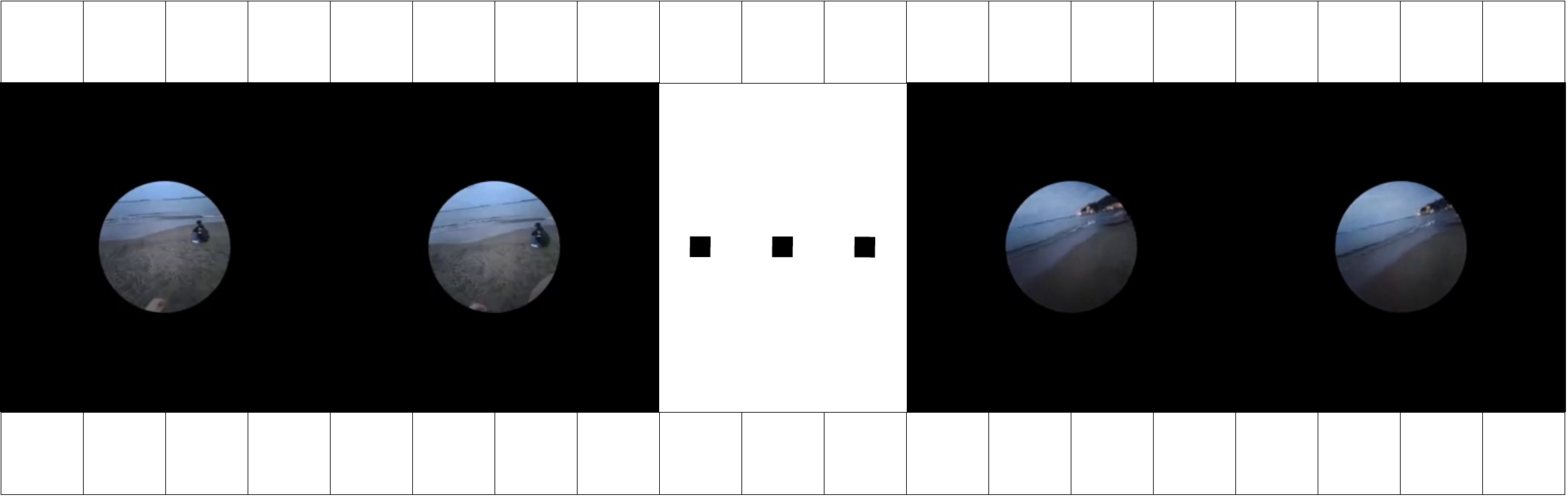}} 
    \caption{Sample videos from the dataset.}
    \label{fig:dataset}
\end{figure}

We conduct our experiments on the 360+x dataset~\cite{chen2024x360}, which was introduced as part of the BinEgo-360 Challenge. The training data are organised into four folders as illustrated in \cref{fig:dataset}. The first contains 360° panoramic videos captured by a static camera. The second provides third-person front-view clips extracted from the panoramas. The third includes egocentric monocular recordings. The fourth contains egocentric binocular clips captured by wearable glasses. In total, the training set consists of more than two thousand videos, covering 28 scene categories (15 indoor and 13 outdoor) and annotated with 38 fine-grained action classes. Each video has an average duration of around six minutes, which is much longer than conventional benchmarks, ensuring that multiple actions occur within a single clip. The combination of multiple views and modalities provides rich contextual cues for both scene classification and temporal action localization tasks.

The competition consists of two rounds. In the first round, the test set contains 16 samples. In the extended round, the test set is enlarged to 39 samples, providing a more reliable evaluation of submitted methods. 

The final ranking is based on mean Average Precision (mAP), computed across action classes and multiple IoU thresholds as follows:
\[
\text{Final Score} = \frac{1}{|T|} \sum_{t \in T} \frac{1}{C} \sum_{c=1}^{C} AP_c^{(t)},
\]
where $T = {0.5, 0.75, 0.95}$ is the set of IoU thresholds and $C$ is the number of classes.

\subsection{Setup}
We implemented our method in PyTorch~\cite{paszke2019pytorch}, running all experiments on a single NVIDIA H100 GPU. For the backbone of the TSM, we adopted ResNeXt-101 models with 64$\times$4d and 32$\times$8d cardinality settings~\cite{xie2017aggregated}. The models were trained using stochastic gradient descent (SGD) with a learning rate of 0.001. All input frames were resized to $256 \times 256$ pixels, and a dropout rate of 0.5 was applied during training. For the ensemble step, the weight of each model was determined by its public score on the leaderboard, ensuring that stronger models contributed more to the final prediction. During training, we used the panoramic videos as input, since the other folders did not provide a complete set of samples.

\subsection{Results}

\begin{table}[h]
\centering
\caption{Top five teams in the first round of the challenge.}
\label{tab:round1}
\begin{tabular}{c|cc}
Team                    & Public score & Private score    \\ \hline
Duong Anh Kiet          & 0.67910      & \textbf{0.52941} \\
iAmAbIrD                & 0.57462      & 0.48235          \\
Loric Bobon             & 0.18656      & 0.17647          \\
Varsovia Hb             & 0.18656      & 0.17647          \\
Yani (Student) Ameziane & 0.18656      & 0.17647         
\end{tabular}
\end{table}

We report the leaderboard results of the BinEgo-360 Challenge in \cref{tab:round1} and \cref{tab:round2}. \Cref{tab:round1} shows the top five teams in the first round, while \cref{tab:round2} presents the results from the extended round with a larger test set. Our method consistently ranked first in both phases, achieving the highest private scores among all participants.

\begin{table}[h]
\centering
\caption{Top five teams in the extended round of the challenge.}
\label{tab:round2}
\begin{tabular}{c|cc}
Team           & Public score & Private score    \\ \hline
Duong Anh Kiet & 0.45238      & \textbf{0.56314} \\
iAmAbIrD       & 0.53968      & 0.45934          \\
yoyobar        & 0.34126      & 0.34948          \\
DASH\_SAJA     & 0.28571      & 0.33131          \\
miiicom        & 0.26984      & 0.31747         
\end{tabular}
\end{table}

\subsection{Ablation Studies}
To better understand the contribution of different design choices, we perform a set of ablation studies. \Cref{tab:ablation} reports the results for both the first and extended rounds of the competition. The term \emph{Single} refers to training the model only on the temporal action localization task, while \emph{Multi} denotes our multi-task setting that jointly optimizes for classification and TAL. The notation 32$\times$8d and 64$\times$4d indicates the ResNeXt backbone used~\cite{xie2017aggregated}. The parameter $t$ (in seconds) controls the interval size when partitioning videos into non-overlapping segments, as described in \cref{sec:method}. Finally, the last row corresponds to our ensemble model, where the weights are determined by the public leaderboard scores of individual method.

\begin{table}[h]
\caption{Ablation results on the BinEgo-360 Challenge.}
\label{tab:ablation}
\resizebox{\linewidth}{!}{
\begin{tabular}{c|cc|cc}
\multirow{2}{*}{Method}      & \multicolumn{2}{c|}{First round} & \multicolumn{2}{c}{Extend round} \\ \cline{2-5} 
                             & Public score   & Private score   & Public score   & Private score   \\ \hline
Baseline                     & 0.18656        & 0.17647         & 0.28571        & 0.39013         \\
Single; 32$\times$8d; t=1.0  & 0.17910        & 0.34117         & 0.06349        & 0.16868         \\
Single; 32$\times$8d; t=0.5  & 0.38059        & 0.44705         & 0.16666        & 0.15916         \\
Single; 32$\times$8d; t=0.25 & 0.35074        & 0.38823         & 0.13492        & 0.11851         \\
Single; 64$\times$4d; t=0.5  & 0.47761        & 0.48235         & 0.19047        & 0.14273         \\
Multi; 32$\times$8d; t=0.5   & 0.51492        & 0.49411         & 0.17460        & 0.18771         \\
Multi; 64$\times$4d; t=0.5   & 0.57462        & 0.51764         & 0.18253        & 0.19377         \\ \hline
Ensemble                     & 0.67910        & 0.52941         & 0.44444        & 0.56314        
\end{tabular}}
\end{table}
\section{Conclusion}
\label{sec:conclusion}

In this paper, we presented our winning solution to the BinEgo-360 Challenge at ICCV 2025. Our method extends the Temporal Shift Module (TSM) to temporal action localization by introducing a background label and applying classification over fixed-length intervals. The multi-task framework allows the model to benefit from both scene classification and TAL supervision, while the ensemble step further stabilizes predictions across different backbones and configurations. Experiments on the challenge dataset confirmed that our approach achieved the highest ranking in both competition rounds, outperforming all other participating teams. 

Although we were not able to perform a full comparison with recent state-of-the-art methods due to the limited time frame of the competition, the results demonstrate the competitiveness of our approach within the challenge setting. Furthermore, our experiments focused only on panoramic videos, without exploiting additional modalities such as third-person views, egocentric binocular recordings, or two-channel audio. This highlights that the 360+x dataset provides a rich and diverse resource that can support more comprehensive multi-modal approaches in the future. We believe that further exploration of these modalities will open new opportunities for advancing temporal action localization and scene understanding in complex real-world environments.

{
    \small
    \bibliographystyle{ieeenat_fullname}
    \bibliography{main}
}

\end{document}